\pdfoutput=1

\documentclass[sigconf]{acmart}
\usepackage{enumerate}
\usepackage{amsmath}
\usepackage{amssymb}
\usepackage{graphicx} 
\usepackage{subfigure}
\usepackage{multirow}

\usepackage{stfloats}
\usepackage{cuted}
\usepackage{capt-of}
\usepackage{marvosym}
\usepackage[noend]{algpseudocode}
\usepackage{algorithmicx,algorithm}

\usepackage{bbding}

\AtBeginDocument{%
  \providecommand\BibTeX{{%
    \normalfont B\kern-0.5em{\scshape i\kern-0.25em b}\kern-0.8em\TeX}}}

\acmConference[MM '21] {Proceedings of the 29th ACM Int'l Conference on Multimedia}{October 20--24, 2021}{Virtual Event, China.}
\acmBooktitle{Proceedings of the 29th ACM Int'l Conference on Multimedia (MM '21), Oct. 20--24, 2021, Virtual Event, China}
\acmPrice{15.00}
\acmISBN{978-1-4503-8651-7/21/10}
\acmDOI{10.1145/3474085.3475532}

\acmSubmissionID{1939}

\copyrightyear{2021}
\acmYear{2021}
\setcopyright{acmcopyright}\acmConference[MM '21]{Proceedings of the 29th ACM International Conference on Multimedia}{October 20--24, 2021}{Virtual Event, China}
\acmBooktitle{Proceedings of the 29th ACM International Conference on Multimedia (MM '21), October 20--24, 2021, Virtual Event, China}
\acmPrice{15.00}
\acmDOI{10.1145/3474085.3475532}
\acmISBN{978-1-4503-8651-7/21/10}

\begin{document}

\title{Object-aware Long-short-range Spatial Alignment for Few-Shot Fine-Grained Image Classification}





\author{Yike Wu$^1$, Bo Zhang$^1$, Gang Yu$^2$, Weixi Zhang$^1$, Bin Wang$^1$, Tao Chen$^1$, Jiayuan Fan$^3$}
\affiliation{%
  \institution{$^1$School of Information Science and Technology, Fudan University, Shanghai, China \\
               $^2$Tencent, Shanghai, China \\
               $^3$Academy for Engineering and Technology, Fudan University, Shanghai, China \\
               }
  }
\email{{ykwu19, zhangb18, weixizhang, wangbin, eetchen, jyfan}@fudan.edu.cn, skicy@outlook.com}

\begin{abstract}
The goal of few-shot fine-grained image classification is to recognize rarely seen fine-grained objects in the query set, given only a few samples of this class in the support set. Previous works focus on learning discriminative image features from a limited number of training samples for distinguishing various fine-grained classes, but ignore one important fact that spatial alignment of the discriminative semantic features between the query image with arbitrary changes and the support image, is also critical for computing the semantic similarity between each support-query pair. In this work, we propose an object-aware long-short-range spatial alignment approach, which is composed of a foreground object feature enhancement (FOE) module, a long-range semantic correspondence (LSC) module and a short-range spatial manipulation (SSM) module. The FOE is developed to weaken background disturbance and encourage higher foreground object response. To address the problem of long-range object feature misalignment between support-query image pairs, the LSC is proposed to learn the transferable long-range semantic correspondence by a designed feature similarity metric. Further, the SSM module is developed to refine the transformed support feature after the long-range step to align short-range misaligned features (or local details) with the query features. Extensive experiments have been conducted on four benchmark datasets, and the results show superior performance over most state-of-the-art methods under both 1-shot and 5-shot classification scenarios.

\end{abstract}




\maketitle

\section{Introduction}

\begin{figure}
    \centering
    \includegraphics[width=7.5cm, height=7.5cm]{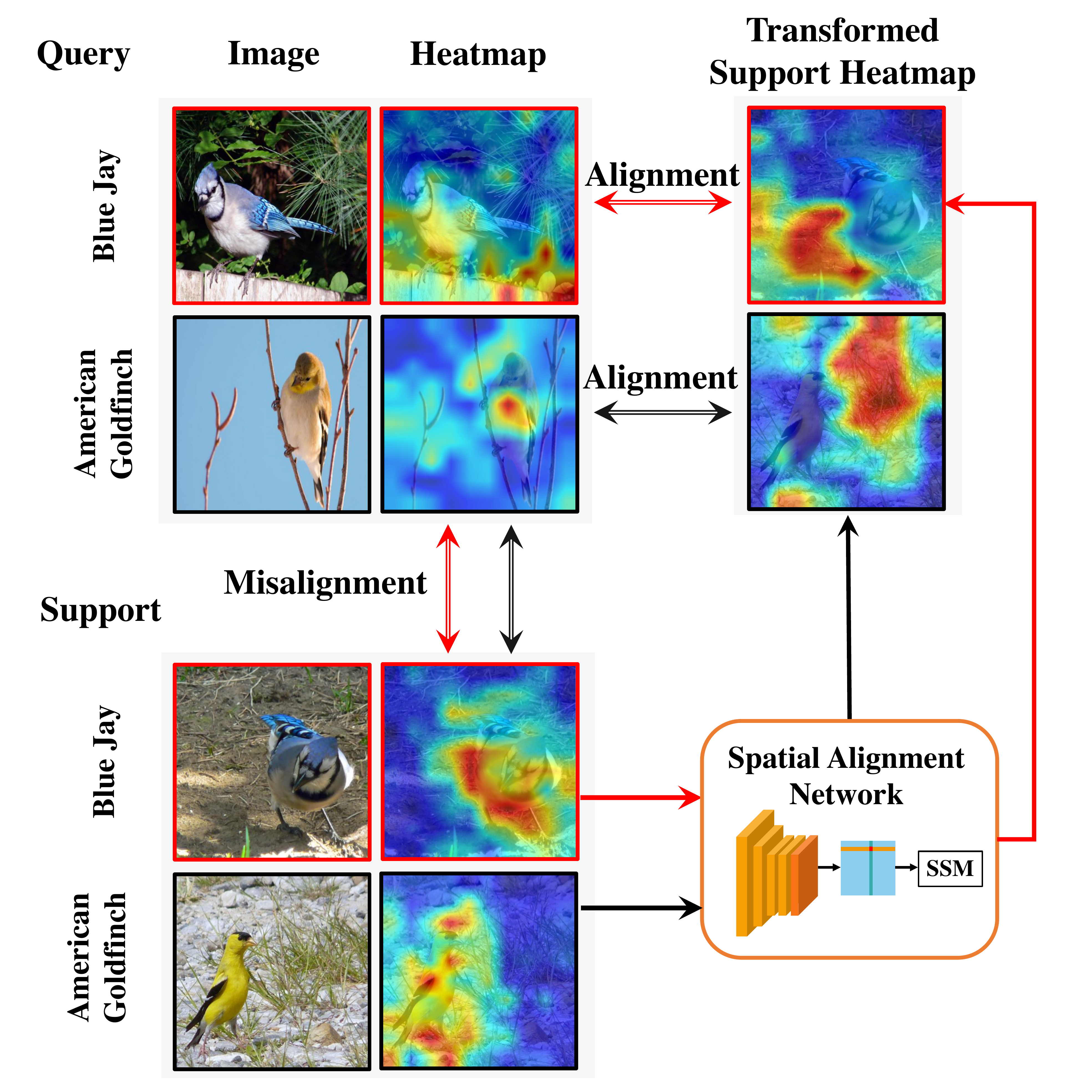}
 \caption{Spatial alignment examples of two different sub-classes from CUB dataset are illustrated, where the red and black color boxes denote Blue Jay and American Goldfinch, respectively. The support and query images in each sub-class are initially misaligned, due to the arbitrary pose and position variations of birds between the two images. By leveraging the proposed spatial alignment network to transform the original support features, the semantic distribution between the transformed support features and the query features can be better aligned. }
 \label{fig1}
\end{figure}

Fine-grained image classification~\cite{lin2015deep, chai2013symbiotic, xie2013hierarchical, zhang2014part, sun2018multi, zheng2017learning, ding2019selective, wang2020graph, lin2015bilinear, gao2016compact, kong2017low, cui2017kernel, zhuang2020learning, du2020fine}, aiming to recognize objects of some sub-classes, plays a critical role in many real-world applications such as the recognition of retail products and the automatic monitoring of biodiversity. Benefiting from the development of Convolution Neural Networks (CNNs)~\cite{lecun1998gradient, simonyan2014very, krizhevsky2017imagenet}, the ability to learn a robust fine-grained prediction model has advanced a lot, on condition that large-scale training samples can be easily accessed and carefully annotated. However, the manual annotation for a large-scale fine-grained image dataset often requires domain-specific knowledge from professionals, which is expensive and difficult to be obtained.

To alleviate the over-dependence of CNNs on large-scale annotated training samples and relieve humans from the cumbersome labeling activities, researchers have explored the meta-learning classification methods for generic categories~\cite{finn2017model, rusu2018meta, lee2019meta, vinyals2016matching, snell2017prototypical, sung2018learning, ye2018learning, zhang2020deepemd, wang2020instance, liu2020negative, lichtenstein2020tafssl, dvornik2020selecting, bateni2020improved}. Specifically, the goal of meta-learning classification is to train a task-agnostic classifier that can be generalized well to different tasks, and to further learn meta-knowledge that can be utilized to compensate for the lack of training data. However, the above meta-learning approaches mainly focus on learning the meta-knowledge for recognizing coarse-grained (generic) categories, which do not have sufficient generalization and representation ability for the fine-grained classification task.

Driven by the success of meta-learning classification models for generic categories, some works ~\cite{li2019distribution, li2019revisiting, zhu2020multi, huang2020low, wei2019piecewise, tang2020revisiting, tsutsui2019meta} try to extend the study of meta-learning from the generic image classification to fine-grained classification, by capturing discriminative parts of the whole image. The work ~\cite{jaderberg2015spatial} tries to globally align images or features via parameterized transformation. Recently, in the work ~\cite{hao2019collect}, a relation matrix is adopted to highlight semantic-related local features. However, these methods fail to achieve fine-grained semantic matching between the samples to be predicted (denoting the query images) and the given training samples (denoting the support images). We argue that accurate spatial alignment for the nuanced feature changes are crucial for fine-grained classification. Thus, how to find the discriminative object features that are useful for fine-grained classification, and further align these discriminative support-query features representing the same semantics but are spatially apart from each other in long or short distance, still require further studies. This can be understood from following aspects.

First, objects often present different scales, poses, etc., with various backgrounds in the images, and in fine-grained image classification the objects play more important roles than backgrounds. We thus need to differentiate the foreground objects from the background, and enhance the discriminative object features that are useful for fine-grained classification, to do further spatial alignment.

Second, ensuring good long-range spatial matching of an object's support-query semantic features is indispensable for the few-shot classification scenario. Since the number of training samples for a class is very small, it is difficult to find a good matching between a limited number of support images and lots of query images with arbitrarily changing object poses, scales and locations, as illustrated in Figure~\ref{fig1}. Therefore, the designed network should be able to transform a support image containing misaligned object semantics with the query image from a long spatial distance range, to make the same object present similar semantic distribution in both the query and support images.

Third, a precise short-range spatial matching is also crucial for the fine-grained classification scenario which is more sensitive to local part changes. This is because there often exist larger intra-class variations than inter-class one for fine-grained classification, due to the local short-range part variations, such as beak rotation of a bird in different images. This phenomenon can be alleviated if the local features of a support object can be spatially aligned with those of a query object.


To this end, we propose an object-aware long-short-range spatial alignment network to keep support-query object semantic consistency, for few-shot fine-grained image classification. The proposed network architecture consists of three sub-network modules. Firstly, a Foreground Object Enhancement (FOE) module is designed to emphasize the discriminative parts of the input images via exploiting deformable convolution and attention extraction. Secondly, an object-level Long-range Semantic Correspondence (LSC) module is designed to transform semantic features of the support image to a new distribution, which is aligned with that of the query image from a long-range distance. In this way, the issue of lacking training images under the few-shot scenario can be alleviated. Specifically, to match semantically correlated regions between each pair of support and query features, a semantic correlation matrix is computed.
A higher value in the matrix indicates semantically more similar regions.
Thirdly, a Short-range Spatial Manipulation (SSM) module is developed to further refine the alignment of local regions of an object under the fine-grained classification. Specifically, the SSM learns the short-range feature coordinate offsets from query features to support features on the basis of the long-range aligned support features.


We conduct extensive experiments on several common benchmarks including the CUB, Stanford Cars, Stanford Dogs and NABirds datasets, and their experimental results demonstrate that the proposed spatial alignment network can bring consistent performance gains for the few-shot fine-grained classification task. Further, we extend this fine-grained classification task to cross-domain scenario, to validate the generalization capability of the proposed approach in the wild.

The main contributions of this paper can be summarized as follows:

\begin{enumerate}[1)]
\item We reveal a crucial aspect to improve the fine-grained classification performance under the few-shot scenario, namely, the long-shot-range spatial alignment between support and query features, which is a pioneer to recognize fine-grained objects.
\item To align semantic features between various query images and a support image, we propose an FOE module and an LSC module for discriminative feature emphasis and object-level semantic distribution alignment, followed by an SSM for local part-level semantic alignment.
\item Extensive experiments on four few-shot fine-grained benchmarks validate that the proposed alignment network brings consistent performance gains under the 1- and 5-shot fine-grained classification scenarios.
\end{enumerate}

\section{Related Works}

\begin{figure*}
  \centering
  \includegraphics[width=17.0cm,height=5.0cm]{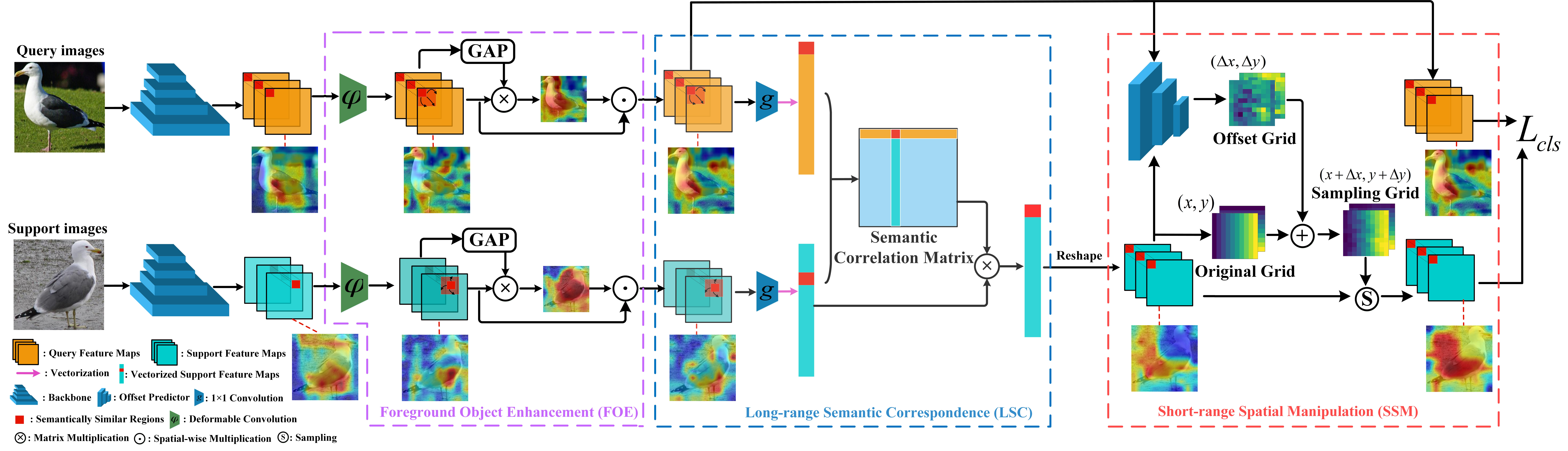}
  \caption{The architecture of our proposed object-aware long-short-range spatial alignment network, consisting of a selected backbone network, a Foreground Object Enhancement (FOE), a Long-range Semantic Correspondence (LSC), and a Short-range Spatial Manipulation (SSM). To clearly illustrate the whole framework, we only visualize a support image having the same object category as the query image. GAP denotes global average pooling.}
  \label{fig2} 
\end{figure*}

\subsection{Few-shot Learning}
Few-shot learning is to learn the modeling of new samples based on a limited number of labeled samples. Most existing works can be roughly categorized into two classes.

In the first class, to achieve quick adaptation to new tasks, most works focus on how to learn a good initialization of parameters. MAML~\cite{finn2017model} first proposes that a model can be adapted to new tasks with only a few gradient descent steps. LEO ~\cite{rusu2018meta} further improves this method via a parameter generative model applied in low-dimensional latent space.
Another solution is MetaOptNet~\cite{lee2019meta}, where the goal is to learn a well-generalized feature embedding of novel classes with a linear support vector machine classifier.

The other class is mainly metric-based methods, which aim to find suitable measurements to quantify the relations between the query images and the given support images. Matching network~\cite{vinyals2016matching} introduces attention mechanism with cosine distance to compute the similarity and designs a Full Context Embedding module. Prototypical Network~\cite{snell2017prototypical} represents each class by calculating the mean features of support images and use Euclidean distance to classify the samples. A deep learnable metric is proposed by Relation Network~\cite{sung2018learning} to replace the previous fixed measurement.
Furthermore, to make full use of the limited number of labeled samples, the semantic alignment approach has been explored in recent works. SAML~\cite{hao2019collect} proposes to align semantic information by highlighting corresponding local regions. ACMM~\cite{huang2019acmm} learns to cross-model align word vectors and pairwise regions. Besides, ~\cite{cao2020learning} proposes to align two self-similarity matrices instead of the relation between the features given two different images. However, these methods try to align features in an one-step way or a global matching way, which is insufficient for fine-grained case where local parts often require minor adjustments.
By comparison, our work aims to mine the semantic correspondence between pair-wise image regions by long-short-range spatial alignment, to transform support features containing misaligned semantics with the query features from both long and short spatial distance ranges.

\subsection{Fine-grained Image Classification}
Fine-grained image classification requires the model to capture the most discriminative features under the interference from various postures of objects and backgrounds, and use these features for image classification.

Early works mainly~\cite{lin2015deep, chai2013symbiotic, xie2013hierarchical, zhang2014part, berg2013poof} rely on various prior information to localize discriminative parts, such as part annotations and bounding boxes, which rely heavily on manual annotations. Other works~\cite{sun2018multi, zheng2017learning, ding2019selective, wang2020graph, lin2015bilinear, gao2016compact, kong2017low, cui2017kernel, jaderberg2015spatial} tend to utilize only image-level annotations to localize discriminative parts. In the work of~\cite{lin2015bilinear}, the high-order feature representation is extracted by a bilinear pooling model. This work is further improved by various feature pooling methods, such as compact bilinear pooling~\cite{gao2016compact}, kernel pooling~\cite{cui2017kernel}, etc. In~\cite{jaderberg2015spatial}, a spatial transformer network is proposed to perform affine transformation to align global features. However, the affine transformation can only perform rotation, shearing, scaling, and translation. This is not enough to handle the huge intra-class differences when the query image and the support image are drastically distinct from each other even if they are of the same sub-class. In this work, we aim to more effectively align the discriminative features for fine-grained classification by a designed spatial alignment network on the enhanced features.

\subsection{Few-shot Fine-grained Image Classification}
Few-shot fine-grained image classification~\cite{wei2019piecewise, zhu2020multi, li2019distribution, li2019revisiting, huang2020low, tsutsui2019meta, zhu2020multi} aims to distinguish the novel sub-classes given a limited number of labeled samples for a generic class. In the work of~\cite{wei2019piecewise}, a bilinear pooling network is proposed to encode features, which are then sent to multiple sub-classifiers to predict the label. In~\cite{zhu2020multi}, a gradient-based meta-learning method is proposed to learn a task-specified initialization for the classifier with multi-attention mechanisms. However, limited effort has been devoted to reducing the effects of local misalignment. The global feature alignment is utilized in~\cite{huang2020low} by learning a transformation matrix to minimize the Euclidean distance between input pairs. More recently, pose normalization alleviates this problem but still requires part annotations during training phase~\cite{tang2020revisiting}. In contrast, we do not need local part annotations for alignment, which relieves experts from expensive labeling and makes our method flexible to use. In our work, we propose a long-short-range spatial alignment network to align support-query semantics for few-shot fine-grained classification.

\section{Method}
\label{sec3}
The overview of the proposed approach is shown in Figure~\ref{fig2}, where both long- and short-range spatial alignment for keeping semantic consistency between the support and query images are imposed. For easy understanding, we first introduce the problem definition of few-shot classification and the common meta-learning baseline models. Next, we give the detailed description of the proposed spatial alignment method. Finally, the overall loss function is given.

\subsection{Preliminaries}
\noindent{\bfseries Problem Definition.}
In the standard setting of few-shot classification, the goal is to learn a generalized network which can be easily adapted to the novel classes $D_n$ with only a few labeled samples, given a large-scale labeled dataset of base classes $D_b$, where $D_{n} \cap D_{b} = \emptyset$. In this work, we study an $N$-way $K$-shot few-shot task, where both the $support$-$set$ $S$ and the $query$-$set$ $Q$ are randomly drawn from the $X_b$ for meta-training, and from the $X_n$ for meta-testing. Besides, the $support$-$set$ $S$ and the $query$-$set$ $Q$ are sampled from the same $N$ classes, with each class containing $K$ labeled samples and $U$ unlabeled samples. Our goal is to assign labels to the unlabeled $N$ $\times$ $U$ samples in the $query$-$set$ $Q$.


\noindent{\bfseries Paired-feature Matching Baseline.}
%
To make use of the support set for matching with the query set, one straightforward meta-leaning baseline model is to directly compare the query features and the given support features in the N-way prediction~\cite{snell2017prototypical}. Specifically, given a pair of images $(I_S, I_Q)$ sampled from the support set $S$ and query set $Q$ respectively, the baseline model first feeds them into the backbone $F$ to extract a pair of high-level features $\left(f_{S}, f_{Q}\right)$, where $f_S = F(I_S)$. Next, a class-wise Euclidean distance between the query features $f_Q$ to be predicted and the $n$-th class support features $f_{n,S}$ is calculated to further represent the inter- and intra-class relationships of the paired features as follows:
\begin{equation}
\begin{aligned}
\label{Meta-training Loss}
L_{cls}(F;S) = &\sum\nolimits_{(I_Q,y_Q)\in Q} log \ P(y_Q = n | I_Q;S) \\&
= \frac{exp(-d(f_{n,S},f_Q))}{\sum_{n^{\prime} \in N} exp(-d(f_{n^{\prime},S}, f_Q))},
\end{aligned}
\end{equation}

\noindent{where $d$($\cdot$, $\cdot$) denotes Euclidean distance between two features, and $y_Q$ is the query label of the corresponding query image $I_Q$. Note that the query label can only be utilized during the meta-training phase, and $f_{n,S}$ denotes support features of the $n$-th class. Besides, $L_{cls}$ is the classification loss function for the query images.}

Considering that postures, scales and local details of objects in fine-grained images vary greatly, it is hard to classify the features of these objects with subtle differences by simply employing the above meta-learning models such as~\cite{sung2018learning, snell2017prototypical, vinyals2016matching}.

Generally, the spatial alignment is addressed by matching paired features \cite{DBLP:conf/cvpr/LeeKPH19, DBLP:journals/tog/LiaoYYHK17} or performing affine transformation~\cite{jaderberg2015spatial}. The former method is a parameter-free way.
The latter method is a learnable way of spatial alignment which is often applied for two images with significant spatial overlapping. In light of these, we design a long-shot-range (global-to-local) spatial alignment network consisting of a Foreground object enhancement (FOE), a Long-range Semantic Correspondence (LSC) and a Short-range Spatial Manipulation (SSM) for each pair of support and query features.



\subsection{Foreground Object Enhancement}
In most scenarios, the features $f \in \mathbb{R}^{C\times H\times W}$ extracted from the images usually present high responses in the object positions, where $C$ is the number of channels, $H$ and $W$ denote the height and width of the features, respectively. However, due to the scattered locations, diverse postures and various scales of the fine-grained objects, it is difficult for the classification network to learn a good mapping between the feature and its real image label. On the other hand, the features $f$ extracted by the backbone network $F$ may not be spatially consistent with the input image. This is caused by the fact that each local feature $f^{x,y} \in \mathbb{R}^{C\times1\times1}$ at the position of $(x, y)$ shares the same receptive field because of the standard convolutions in the backbone network $F$. Consequently, some weakly related or unrelated areas with the object may also be activated, which will interfere the object feature's spatial alignment and subsequent feature comparisons.

Therefore, to refine the feature $f^{x,y}$ to weaken background disturbance and encourage higher foreground object response, we develop a foreground object enhancement (FOE) module and attach the module after the backbone network $F$, which consists of a deformable convolution block $\varphi$ and a spatial attention block. Particularly, the deformable convolution block with learnable parameters $\varphi$ predicts offset sampling positions, to model geometric transformation. In this way, an input-dependent receptive field can be generated to fit the fine-grained objects more precisely. The refined feature is denoted as $\varphi(f)$.

In order to fully suppress the noisy background while emphasizing the foreground objects for both support and query features, we use a soft attention mask $\mathbf{MA}$ as a weight-map of the features.

In detail, given refined features $\varphi(f) \in \mathbb{R}^{C\times H\times W}$, we first perform the global average pooling (GAP) operation for the $\varphi(f)$ to obtain the global semantic representations $u \in \mathbb{R}^{C\times1\times1}$. Next, the soft mask $\mathbf{MA} \in \mathbb{R}^{1\times H\times W}$ is obtained by calculating the cosine similarities between the global semantics $u \in \mathbb{R}^{C\times1\times1}$ and local feature $\varphi(f)^{x,y} \in \mathbb{R}^{C\times1\times1}$ at the local position of $(x,y)$, which can approximately reflect the spatial distribution of the discriminative semantic parts as follows:
\begin{equation}
\label{Soft mask}
\mathbf{MA}_{x,y} = \left\langle\|u\|_{2} \ , \ \|\varphi(f)^{x,y}\|_{2}\right\rangle,
\end{equation}
where $\langle\cdot, \cdot\rangle$ denotes dot product between the two vectors,  $\| \cdot \|$ is the $\mathcal{L}_2$ normalization, and $\mathbf{MA}_{x,y}$ represents the semantic correlations between global semantics and each local feature at the position $(x,y)$. Besides,
due to that the values of the soft mask obtained via Eq. \ref{Soft mask} are within a range of $\mathbf{MA} \in[-1,1]^{H \times W}$, we calculate the normalized mask by $\mathbf{MA} = (\mathbf{MA} + 1) / 2$. Then, the output features ${\dot{f}} \in \mathbb{R}^{C \times H  \times W}$ of the FOE is formulated as follows:
\begin{equation}
\label{Reweighted feature}
{\dot{f}} = \mathbf{MA} \odot \varphi(f),
\end{equation}

\noindent where $\odot$ indicates the spatial-wise multiplication.

\subsection{Long-range Semantic Correspondence}

Given the re-weighted features $\dot{f}$ above, we propose to spatially transform the support features $\dot{f_{S}}$ to align with the query features $\dot{f_{Q}}$ especially for those semantic-related regions.
In order to map the features into low dimensional space, we use $g$ with kernel size of $1 \times 1$ to reduce the dimensions of the above support features and query features.
Next, the semantic correlation matrix $MT$ $\in$ $\mathbb{R}^{HW \times HW}$ is obtained as follows:
\begin{equation}
\label{Correlation matrix}
MT_{i,j} = \frac{\langle \Vec{f_{Q}^i} \ , \ \ \Vec{f_{S}^j \rangle}}{|| \ \Vec{f_{Q}^i}|| \ \ ||\Vec{f_{S}^j}||},
\end{equation}

\noindent where $\Vec{f_{Q}} \in \mathbb{R}^{C \times HW} $ and $\Vec{f_{S}} \in \mathbb{R}^{C \times HW}$ represent the spatially vectorized features of the re-weighted support features $\dot{f_{S}}$ and query features $\dot{f_{Q}}$, respectively.

Given the semantic correlation matrix $MT_{i,j}$, solving the long-range semantic correspondence is simplified as performing spatial alignment for the highly correlated positions in the matrix $MT_{i,j}$. We treat this as a global sampling problem supposing that the support features $\Vec{f_{S}}$ can be spatially rearranged to match the query features $\Vec{f_{Q}}$. Specifically, each row in the semantic correlation matrix $MT_{i,j}$ should be normalized to sum to 1, in order to be used as a weighting vector for the support features $\Vec{f_{S}}$ as follows:
\begin{small}
\begin{equation}
\label{Linear Normlization}
\overline{MT}_{i,j} = \frac{exp(MT_{i,j})}{\sum_{k=1}^{HW} \ exp(MT_{k,j})},
\end{equation}
\end{small}

\noindent where $\overline{MT}_{i,j}$ denotes the row-wise normalized semantic correlation matrix.

Further, the spatially transformed support features $\overline{f}_{S} \in \mathbb{R}^{C \times HW}$ is obtained by calculating the matrix multiplication between the row-wise normalized semantic correlation matrix $\overline{MT}$ and the transpose of vectorized support features $\Vec{f_{S}}$ as follows:
\begin{small}
\begin{equation}
\label{Correlation matrix}
\overline{f}_S = ( \overline{MT} \ \Vec{f_{S}}^T )^T.
\end{equation}
\end{small}

\subsection{Short-range Spatial Manipulation}
Globally aligning features belonging to the same class via the above LSC is beneficial to class-agnostic feature variations such as object postures and positions. However,
some short-range feature variations of local details for fine-grained objects, such as the shape and rotation of a bird beak, being crucial for sub-category bird recognition, cannot be modeled only by the LSC. Accordingly, we further develop a short-range spatial manipulation module to align those nuanced parts of the paired support and query samples for feature refinement, on the basis of well-aligned long-range features. Our idea is to predict a coordinate offset for each local feature $\overline{f}_{S}^{x,y} \in \mathbb{R}^{C \times 1 \times 1}$ at the original position $(x,y)$.

Specifically, the SSM is composed of three stages: a sampling grid, a coordinate offset predictor $h$, and a differentiable feature sampling.

First, given the well transformed features $\overline{f}_{S}$ after LSC, we first generate the original coordinate grid, whose values are within a range of $[-1, 1]^{H \times W \times 2}$, where the number 2 denotes two channels, representing x- and y-coordinates, respectively. The left-top coordinate value of the grid is denoted as $(-1, -1)$ while the corresponding coordinate value in the right bottom is denoted as $(1, 1)$. Note that the local feature $\overline{f}_{S}^{x,y}$ at the same position $(x,y)$ of different channels shares the same coordinate.

Next, we concatenate the transformed support features $\overline{f}_S$ and the query features $\dot{f_{Q}}$, and feed the concatenated features into the subsequent offset predictor. In this way, the offset predictor can generate an offset grid, representing the position offsets $(\Delta x, \Delta y)$.
For each initial coordinate $(x,y)$ of the support feature $\overline{f}_{S}^{x,y}$, we calculate its corresponding rectified sampling grid $(\hat{x}, \hat{y})$ as follows:
\begin{equation}
\begin{aligned}
\label{sampling grid}
\left( \begin{array}{c}
    \hat{x}\\
    \hat{y}\\
  \end{array}
  \right) =  \left( \begin{array}{c}
    x\\
    y\\
  \end{array}
  \right) +  \left( \begin{array}{c}
    \Delta{x} \\
    \Delta{y} \\
  \end{array}
  \right),
\end{aligned}
\end{equation}

\noindent where $\Delta x=h([\overline{f}_S, \ \dot{f_{Q}}])$ denotes the offset predictor with learnable parameters $h$, and $[\cdot,\cdot]$ is the concatenation operation.

Finally, the sampled value of the local feature $\hat{f}_S^{\hat{x}, \hat{y}}$ at the rectified position of ($\hat{x}$, $\hat{y}$) can be obtained by calculating the distance-weighted sum of its four neighboring features as follows:
\begin{equation}
\begin{aligned}
\label{Correlation matrix}
\hat{f}_S^{\hat{x}, \hat{y}} =\sum_n^H \sum_m^W & (\overline{f}_S^{n, m} \ max(0, 1-|\hat{x}-m|) \ max(0, 1-|\hat{y}-n|)).
\end{aligned}
\end{equation}

\subsection{Overall Loss Function}

In the meta-training stage, the task is to learn a task-agnostic backbone network $F$ serving as the classification network. Specifically, an N-way K-shot task will be randomly sampled from a task distribution $P(\mathcal{T})$ in each episode $T$. Given N$\times$1 labeled images from the support set $S$, the final network tries to predict the sub-category label for unlabeled query images from the query set $Q$. Based on the selected backbone network $F$ and well-designed $\varphi$, $g$, and $h$, the overall meta-loss considering the long-shot-range spatial alignment can be defined as follows:
\begin{equation}
\begin{aligned}
\label{Meta-training Loss}
L_{cls}(F,\varphi, g, h;S) = &\sum\nolimits_{(I_Q,y_Q)\in Q} log \ P(y_Q=n|I_Q;S) \\& = \frac{exp(-d(h(g(\varphi(f_{n,S}))),f_Q))}{\sum_{n^{\prime} \in N} exp(-d(h(g(\varphi(f_{{n^{\prime},S}}))), f_Q))}.
\end{aligned}
\end{equation}

\section{Experiments}
We conduct our experiments on four fine-grained benchmark datasets including CUB, Stanford Dogs, Stanford Cars and NABirds. All experiments of the proposed method are implemented by Pytorch, and all images are resized to $84\times84$ pixels.

\vspace{-0.10cm}
\subsection{Dataset}

\noindent{\bfseries CUB}
~\cite{wah2011caltech} is the most widely used fine-grained dataset, containing 11,788 images collected from 200 bird categories. We follow the same split used in ~\cite{zhang2020deepemd}, which is divided into 100, 50 and 50 classes for meta-train, meta-validation and meta-test, respectively.

\noindent{\bfseries Stanford Dogs}
~\cite{khosla2011novel} includes 120 sub-classes of dogs and 20,580 images in total. Following~\cite{zhu2020multi}, we adopt 70, 20, 30 classes for meta-train, meta-validation and meta-test, respectively.

\noindent{\bfseries Stanford Cars}
~\cite{krause20133d} contains 16,185 images coming from 196 sub-classes. Following ~\cite{zhu2020multi}, we adopt 130, 17, 49 classes for meta-train, meta-validation and meta-test, respectively.

\noindent{\bfseries NAB}
~\cite{van2015building} contains 555 sub-classes of the North American bird. The split is consistent with~\cite{huang2020low}, which uses 350, 66, 139 categories for meta-train, meta-validation and meta-test, respectively.

\begin{table*}[]
\setlength{\tabcolsep}{2mm}{
\begin{tabular}{cccccccc}
\hline
\multirow{2}{*}{Method} & \multirow{2}{*}{Backbone} & \multicolumn{2}{c}{Stanford Dogs}         & \multicolumn{2}{c}{Stanford Cars}         & \multicolumn{2}{c}{NABirds}               \\
                        &                           & 1-shot              & 5-shot              & 1-shot              & 5-shot              & 1-shot              & 5-shot              \\ \hline
Matching Net$^\dagger$~(NIPS-16)~\cite{vinyals2016matching}            & Conv-64                   & 35.80±0.99          & 47.50±1.03        & 34.80±0.98        & 44.70±1.03        & -                   & -                   \\
Prototypical Net$^\dagger$~(NIPS-17)~\cite{snell2017prototypical}        & Conv-64                   & 37.59±1.00        & 48.19±1.03        & 40.90±1.01        & 52.93 ± 1.03        & -                   & -                   \\
RelationNet~(CVPR-18)~\cite{sung2018learning}             & Conv-64                   & 43.29±0.46$^\diamond$          & 55.15±0.39$^\diamond$          & 47.79±0.49$^\diamond$          & 60.60±0.41$^\diamond$          & 64.34 ±0.81*          & 77.52±0.60*          \\
GNN$^\dagger$~(ICLR-18)~\cite{garcia2017few}                     & Conv-64                   & 46.98±0.98        & 62.27±0.95        & 55.85±0.97        & 71.25±0.89        & -                   & -                   \\
CovaMNet~(AAAI-19)~\cite{li2019distribution}                & Conv-64                   & 49.10±0.76          & 63.04±0.65          & 56.65±0.86          & 71.33±0.62          & 60.03±0.98*          & 75.63±0.79*          \\
DN4~(CVPR-19)~\cite{li2019revisiting}                     & Conv-64                   & 45.73±0.76          & 66.33±0.66          & 61.51±0.85          & \textbf{89.60±0.44} & 51.81±0.91*          & 83.38±0.60*          \\
LRPABN~(TMM-20)~\cite{huang2020low}                  & Conv-64                   & 45.72±0.75          & 60.94±0.66          & 60.28±0.76          & 73.29±0.58          & 67.73±0.81*          & 81.62±0.58*          \\
MattML~(IJCAI-20)~\cite{zhu2020multi}                  & Conv-64                   & 54.84±0.53          & 71.34±0.38          & 66.11±0.54          & 82.80±0.28          & -                   & -                   \\
ATL-Net~(IJCAI-20)~\cite{ijcai2020-100}                  & Conv-64                   & 54.49±0.92          & \textbf{73.20±0.69}          & 67.95±0.84          & 89.16±0.48          & -                   & -                   \\\hline
Ours                    & Conv-64                   & \textbf{55.53±0.45} & 71.68±0.36 & \textbf{70.13±0.48} & 84.29±0.31          & \textbf{75.60±0.49} & \textbf{87.21±0.29} \\
Ours                    & ResNet-12                 & \textbf{64.15±0.49} & \textbf{78.28±0.32} & \textbf{77.03±0.46} & 88.85±0.46          & \textbf{83.76±0.44} & \textbf{92.61±0.23} \\ \hline
\end{tabular}}
\caption{\label{tab1}5-way classification accuracies ($\%$) on the Stanford Dogs, Stanford Cars and NABirds datasets respectively. The $\pm$ denotes that the results are reported with 95$\%$ confidence intervals over 2000 test episodes. $^\dagger$: reported in ~\cite{li2019distribution}. $^\diamond$: reported in ~\cite{zhu2020multi}. * represents that the results are reported in~\cite{huang2020low}. Other results are reported in the original papers.}
\end{table*}

\begin{table}[]
\setlength{\tabcolsep}{0.41mm}{
\begin{tabular}{cccc}
\hline
\multirow{2}{*}{Method} & \multirow{2}{*}{Backbone} & \multicolumn{2}{c}{CUB}                   \\
                        &                           & 1-shot              & 5-shot              \\ \hline
Matching Net$^\circ$~(NIPS-16)~\cite{vinyals2016matching}              & Conv-64                   & 67.73±0.23          & 79.00±0.16          \\
ProtoNet$^\circ$~(NIPS-17)~\cite{snell2017prototypical}               & Conv-64                   & 63.73±0.22          & 81.50±0.15          \\
FEAT~(CVPR-20)~\cite{ye2018learning}                    & Conv-64                   & 68.87±0.22          & 82.90±0.15          \\ \hline
Ours                    & Conv-64                   & \textbf{73.07±0.46} & \textbf{86.24±0.29} \\ \hline
ProtoNet*~(NIPS-17~\cite{snell2017prototypical}               & ResNet-12                 & 66.09±0.92          & 82.50±0.58          \\
RelationNet*~(CVPR-18)~\cite{sung2018learning}             & ResNet-34                 & 66.20±0.99          & 82.30±0.58          \\
MAML*~(ICML-17)~\cite{finn2017model}                    & ResNet-34                 & 67.28±1.08          & 83.47±0.59          \\
cosine classifier*~(ICLR-17)~\cite{chen2019closer}      & ResNet-12                 & 67.30±0.86          & 84.75±0.60          \\
Matching Net*~(NIPS-16)~\cite{vinyals2016matching}              & ResNet-12                 & 71.87±0.85          & 85.08±0.57          \\
DeepEMD~(CVPR-20)~\cite{zhang2020deepemd}                & ResNet-12                 & 75.65±0.83          & 88.69±0.50          \\
ICI~(CVPR-20)~\cite{wang2020instance}                     & ResNet-12                 & 76.16               & \textbf{90.32}               \\ \hline
Ours                    & ResNet-12                 & \textbf{77.77±0.44} & 89.87±0.24 \\ \hline
\end{tabular}}
\caption{\label{tab_cub}5-way classification accuracies ($\%$) on the CUB dataset. The definition of $\pm$ follows Table~\ref{tab1}. $^\circ$ denotes the results are reported in ~\cite{ye2018learning}. *: reported in ~\cite{zhang2020deepemd}.}
\end{table}

\vspace{-0.10cm}
\subsection{Network Structure}
To make a fair comparison with prior works, we adopt the commonly used Conv-64~\cite{snell2017prototypical, ye2018learning} and ResNet-12~\cite{zhang2020deepemd, chen2020new, oreshkin2018tadam} as the backbone network $F$, where the size of the output features $f$ is $64\times 10 \times 10$ and $512\times 10 \times 10$, respectively.
Note that we remove the last pooling layer of the backbone network in order to preserve the spatial resolution of features, and no additional parameters are introduced.

\noindent{The} operation $g$ used in LSC contains a $1 \times 1$ convolutional layer, a BatchNorm layer, a ReLU layer and a $1 \times 1$ convolutional layer.

\noindent{The offset predictor $h$} used in SSM consists of three convolutional blocks. The first two blocks each consists of a convolutional layer with kernel size of $3 \times 3$, a BatchNorm layer, and a ReLU layer. The last block contains a $3 \times 3$ convolutional layer and a Tanh layer.

\vspace{-0.10cm}
\subsection{Experimental Setup}

\noindent{\bfseries Pre-training Phase.}
Following the pre-training setting in~\cite{chen2020new}, we insert a fully-connected layer at the end of the backbone network, and optimize the network on base classes via minimizing the cross-entropy loss. Standard data augmentation including random crop and horizontal flipping is employed. For all datasets, we train 200 epochs with a batch size of 128. SGD optimizer is used with a learning rate of 0.1, a decay factor of 0.1, and the learning rate decays at 85 and 170 epochs. The weight decay for Conv-64 and ResNet-12 is 0.0005. After the pre-training phase, we remove the fully-connected layer for the next two meta-training phases.

\noindent{\bfseries Episodic Meta-training LSC Phase.}
For all three datasets, we train 200 epochs with 200 batches per epoch for both 1- and 5-shot settings. Each batch contains 4 episodes, and each episode includes 5 classes with 15 query images per class. Adam optimizer with a fixed learning rate of 0.001 is applied for CUB, NABirds, and Stanford Cars, and 0.0005 for Stanford Dogs, respectively. Data augmentation methods such as random crop, horizontal flip, and color jittering are used. Moreover, the best model is selected according to its classification accuracy on the meta-validation set.

\noindent{\bfseries Episodic Meta-training SSM Phase.}
In this phase, we follow the settings of data augmentation and model selection in the above meta-training LSC phase. Besides, Adam optimizer is used with a fixed learning rate of 0.0001 for Stanford Cars, NABirds and CUB, 0.00005 for Stanford Dogs, respectively. 50 epochs are trained for all datasets.

\vspace{-0.10cm}
\subsection{Experimental Results}

\noindent{\bfseries Stanford Dogs, Stanford Cars and NABirds.}
 Table~\ref{tab1} reports the 5-way classification results on the Stanford Dogs, Stanford Cars and NABirds, respectively. During the meta-testing phase, the evaluation results are calculated with 95\% confidence intervals over 2000 test episodes.

 It can be seen that the proposed method achieves the best results under the most settings on three datasets. Specifically, our method outperforms state-of-the-art few-shot fine-grained classification methods by $0.69\%$, $4.02\%$, $7.87\%$ under 1-shot learning on Stanford Dogs, Stanford Cars and NABirds respectively. Besides, we conduct experiments on another backbone, ResNet-12, further showing the generalization of our methods.

 NABirds is more challenging than CUB due to its diverse distribution of categories. To further show our method's generalization ability in fine-grained classification, we conduct the experiments on NABirds following the setup in~\cite{huang2020low}. As shown in Table~\ref{tab1}, the proposed method also achieves the highest classification accuracy over the existing methods.


 \noindent{\bfseries CUB.}
 Considering that the existing state-of-the-art few-shot learning methods also report the results on CUB, we carry out experiments following the same setting used in ~\cite{ye2018learning}. Note that all compared methods use the images cropped by bounding boxes. In Table~\ref{tab_cub}, it can be observed that the proposed method achieves superior classification accuracy in most scenarios, showing the advantages of the proposed method under data-scarce scenarios.

\subsection{Insight Analyses}
\begin{table}[]
\begin{tabular}{cccccc}
\hline
\multirow{2}{*}{Baseline} & \multirow{2}{*}{FOE} & \multirow{2}{*}{LSC} & \multirow{2}{*}{SSM} & CUB                 & Stanford Cars       \\
                          &                      &                      &                      & 1-shot              & 1-shot              \\ \hline
$\checkmark$                          &                      &                      &                      & 63.89±0.49          & 50.67±0.44          \\
$\checkmark$                          &                      & $\checkmark$                     &                      & 69.22±0.51          & 60.62±0.49          \\
$\checkmark$                          & $\checkmark$                     &                      &                      & 69.50±0.49          & 64.57±0.49          \\
$\checkmark$                          &                      & $\checkmark$                     & $\checkmark$                     & 70.80±0.51          & 62.83±0.48          \\
$\checkmark$                          & $\checkmark$                     & $\checkmark$                     &                      & 71.35±0.47          & 67.99±0.49          \\
$\checkmark$                          & $\checkmark$                     & $\checkmark$                     & $\checkmark$                     & \textbf{73.07±0.46} & \textbf{70.13±0.48} \\ \hline
\end{tabular}
\caption{\label{tab_ablation} Ablation studies of each module on CUB and Stanford Cars 5-way 1-shot classification task. Conv-64 is used as the backbone. The definition of $\pm$ follows Table~\ref{tab1}.}
\end{table}

\begin{table}[]
\begin{tabular}{cccccc}
\hline
\multirow{2}{*}{Baseline} & \multicolumn{2}{c}{FOE} & CUB                 & Dogs                & Cars                \\
                          & $\varphi$    & $\mathbf{MA}$   & 1-shot              & 1-shot              & 1-shot              \\ \hline
$\checkmark$                          &           &             & 63.89±0.49          & 45.92±0.44          & 50.67±0.44          \\
$\checkmark$                          &           & $\checkmark$            & 63.85±0.48          & 46.48±0.44          & 50.64±0.45          \\
$\checkmark$                          & $\checkmark$          &             & 69.19±0.49          & 50.28±0.45          & 64.31±0.48          \\
$\checkmark$                          & $\checkmark$          & $\checkmark$            & \textbf{69.50±0.49} & \textbf{50.40±0.43} & \textbf{64.57±0.49} \\ \hline
\end{tabular}
\caption{\label{tab_FOE} Ablation studies of FOE on three datasets on 5-way 1-shot classification task. Conv-64 is used as the backbone. The definition of $\pm$ follows Table~\ref{tab1}.}
\end{table}

\begin{table}[]
\setlength{\tabcolsep}{1.2mm}{
\begin{tabular}{ccccccc}
\hline
\multirow{2}{*}{Baseline} & \multirow{2}{*}{FOE} & \multicolumn{2}{c}{LSC} & CUB        & Dogs       & Cars       \\
                          &                      & $g$        & $\overline{MT}$       & 1-shot     & 1-shot     & 1-shot     \\ \hline
$\checkmark$                          &                      &          &              & 63.89±0.49 & 45.92±0.44 & 50.67±0.44 \\
$\checkmark$                          &                      &          & $\checkmark$             & 67.61±0.51 & 50.39±0.49 & 59.85±0.50 \\
$\checkmark$                          &                      & $\checkmark$         & $\checkmark$             & 69.22±0.51 & 50.39±0.49 & 60.62±0.49 \\
$\checkmark$                          & $\checkmark$                     &          &              & 69.50±0.49 & 50.40±0.43 & 64.57±0.49 \\
$\checkmark$                          & $\checkmark$                     &          & $\checkmark$             & 71.24±0.48 & 53.18±0.45 & 67.83±0.48 \\
$\checkmark$                          & $\checkmark$                     & $\checkmark$         & $\checkmark$             & \textbf{71.35±0.47} & \textbf{55.06±0.46} & \textbf{67.99±0.49} \\ \hline
\end{tabular}}
\caption{\label{tab_LSC} Ablation studies of LSC on three datasets on 5-way 1-shot classification task. Conv-64 is used as the backbone. The definition of $\pm$ follows Table~\ref{tab1}.}
\end{table}

\begin{table}[]
\setlength{\tabcolsep}{4mm}{
\begin{tabular}{cccc}
\hline
\multirow{2}{*}{Method} & \multirow{2}{*}{Backbone} & \multicolumn{2}{c}{CUB $\to$ NABirds} \\
                        &                           & 1-shot                 & 5-shot                 \\ \hline
Baseline                & ResNet-12                 & 45.70±0.45             & 63.84±0.40             \\
Ours                    & ResNet-12                 & \textbf{48.50±0.48}    & \textbf{66.35±0.41}                        \\ \hline
\end{tabular}
\caption{\label{tab_cross}Cross-domain adaptation from the CUB to NABirds dataset. The results are reported on 5-way classification accuracies ($\%$). The definition of ± follows Table~\ref{tab1}.}}
\end{table}

\noindent{\bfseries Module-wise Ablation Study.}
In this section, we investigate the impact of individual module. The overall results are presented in Table ~\ref{tab_ablation}. Exciting accuracy gains come from the following aspects. Firstly, LSC significantly increases the accuracy of the baseline model from $63.89\%$ to $69.22\%$ on CUB. Secondly, by introducing FOE for object feature enhancement, the accuracy of the baseline model can be increased. Thirdly, SSM can further boost the accuracy to a higher level.

\noindent{\bfseries Ablation Study of FOE.}
We evaluate the impact of $\varphi$ and $\mathbf{MA}$ employed in FOE on three datasets. Table ~\ref{tab_FOE} presents that each component brings performance gains due to that the feature learning is focused on the foreground object. Besides. the most significant gain comes from $\varphi$, $5.3\%$, $4.5\%$, $13.64\%$ for CUB, Stanford Dogs and Stanford Cars, respectively. Further, we observe that $\mathbf{MA}$ can further improve the results by suppressing semantically irrelevant regions.

\begin{figure*}[htbp]
    \centering
    \includegraphics[scale=0.34]{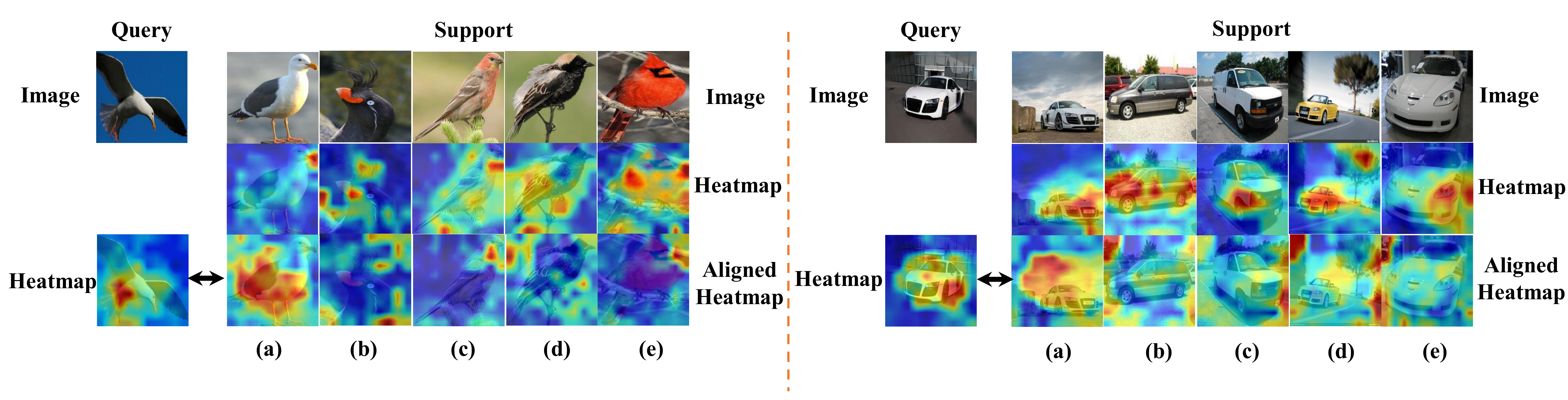}
 \caption{Visualization results of features after the LSC module on CUB and Stanford Cars. Note that two datasets show the same situation, where the support image (a) is of the same class and (b) $\sim$ (e) are of different classes with the query image, respectively. LSC can produce aligned support features having highest similarities in the same class.}
 \label{lsc}
  \vspace{-0.3cm}
\end{figure*}

\begin{figure}
    \centering
    \includegraphics[scale=0.28]{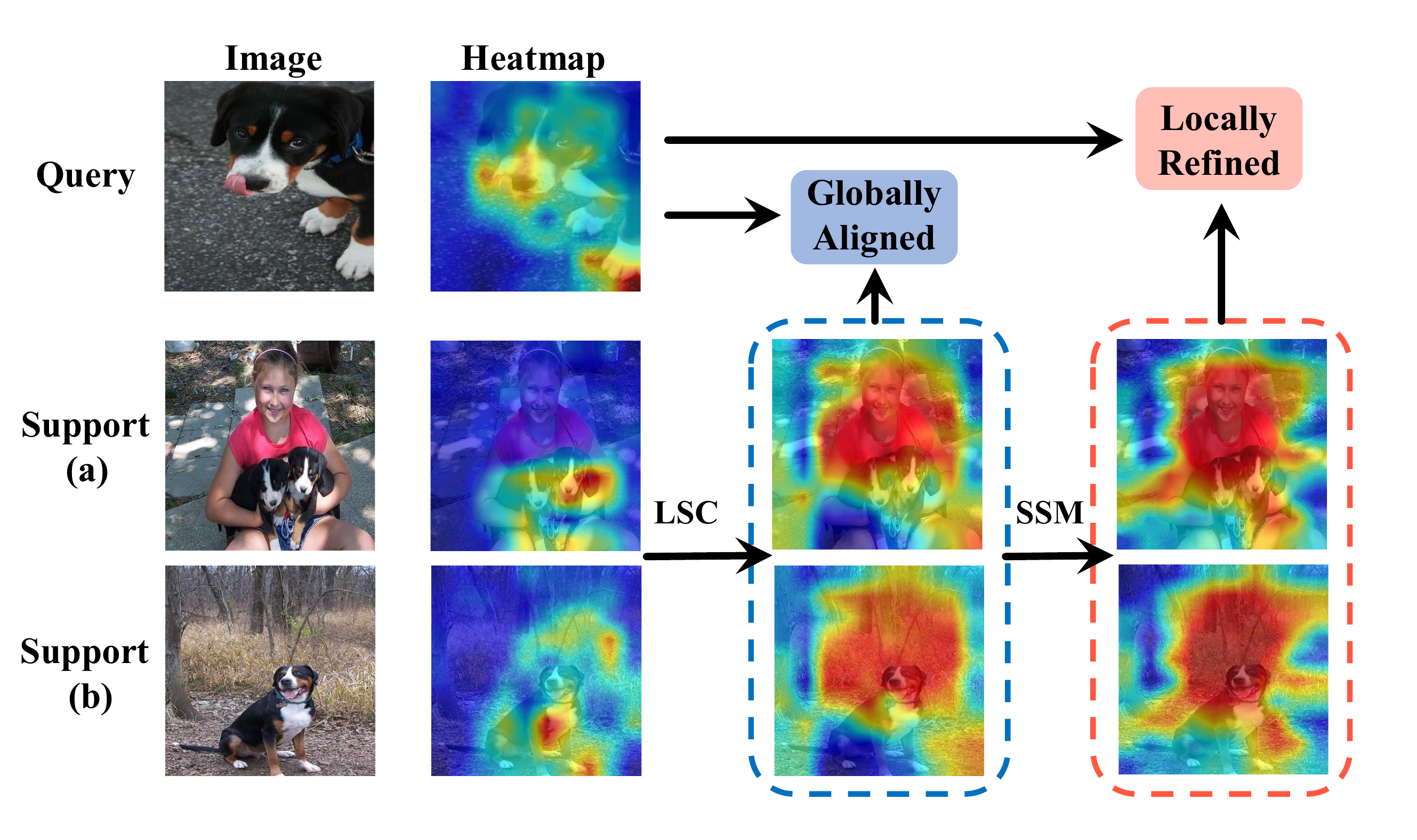}
 \caption{Visualization results of features after the LSC module and SSM module on Stanford Dogs dataset. Support Image (a), (b) and the query image are of the same class.}
 \label{ssm}
 \vspace{-0.3cm}
\end{figure}

\begin{figure}
    \centering
    \includegraphics[scale=0.25]{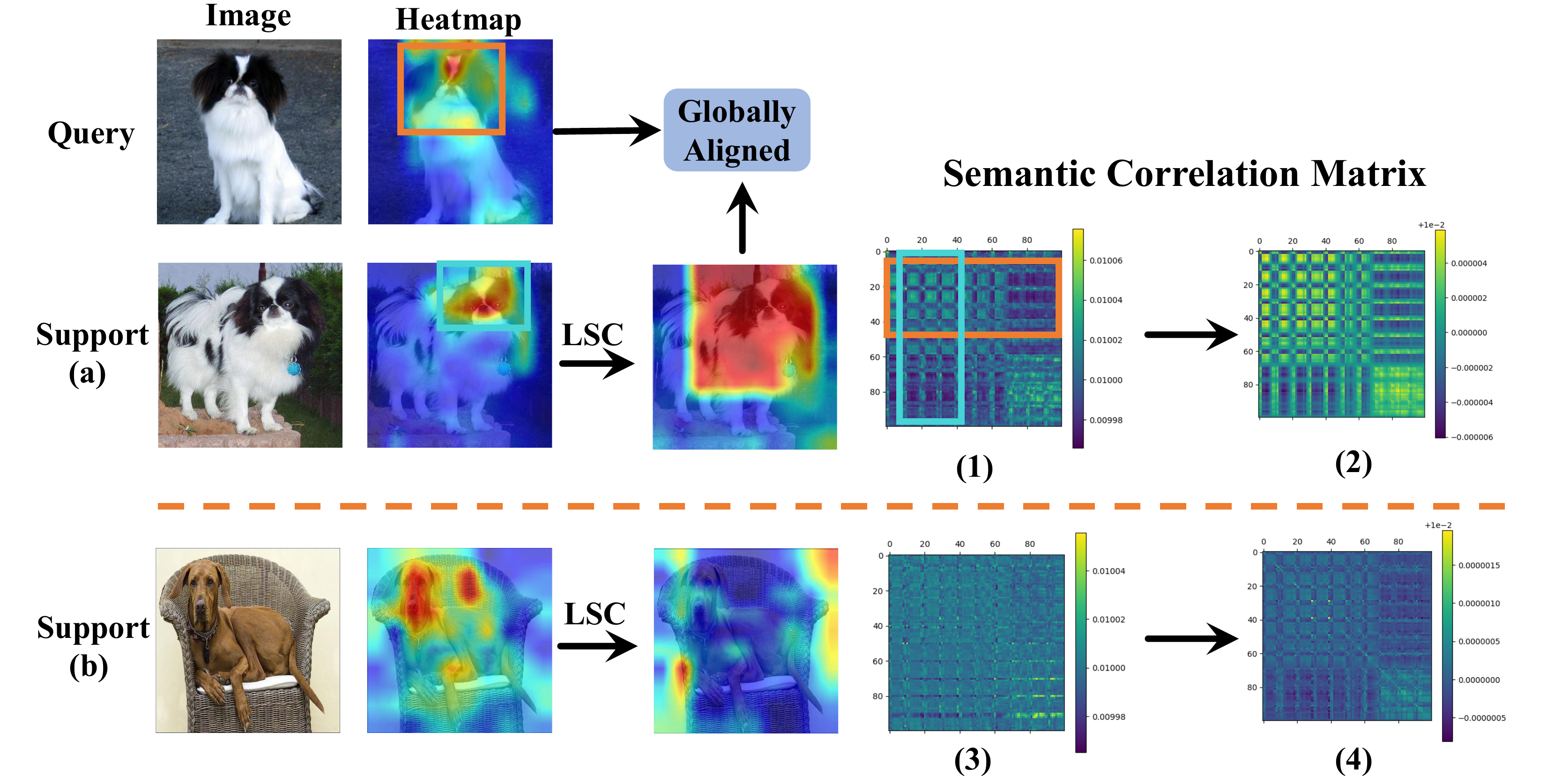}
 \caption{Visualization results of features after the LSC module on Stanford Dogs dataset. The query image and support image (a) are of the same class, support image (b) is of different class. The visualized matrices (1) and (3) represent the semantic correlation matrices before the LSC module. The matrices (2) and (4) are the visualized semantic correlation matrices after the LSC module.}
 \label{matrix}
 \vspace{-0.3cm}
\end{figure}

\noindent{\bfseries Ablation Study of LSC.}
To further study the impact of $g$ and $\overline{MT}$ in LSC, we conduct experiments in Table ~\ref{tab_LSC}. The addition of $g$ brings further improvement to the results. Considering that certain gains are due to FOE as discussed above, we report the results when only LSC is employed on three datasets. It can be seen that LSC brings significant improvements to the baseline model, by $5.33\%$, $4.47\%$, $9.95\%$ for CUB, Stanford Dogs and Stanford Cars, respectively. Also, LSC can further boost the accuracy after introducing FOE, which shows the effectiveness of the proposed spatial alignment.

\noindent{\bfseries Cross-domain Fine-grained Classification.} We further extend the study of few-shot fine-grained classification to a real-world application. Considering such an important fine-grained recognition scenario: given generic bird categories widely collected from the Internet, the goal of cross-domain fine-grained classification is to adapt the well-trained fine-grained classifier trained on the generic bird categories to specific birds living in a particular location such as the North American.
To our best knowledge, this is the first attempt that the cross-domain fine-grained classification is studied under the few-shot scenario. It can be seen from Table~\ref{tab_cross} that
the proposed 
method is beneficial to recognize rarely seen species of birds, which is helpful for the real-world applications.

\subsection{Visualization Results}
To demonstrate that the proposed spatial alignment network is beneficial for the baseline model, we visualize the support and query features, and the corresponding aligned support features. Specifically, in Figures~\ref{lsc}, ~\ref{ssm} and ~\ref{matrix}, the heatmaps are obtained via a max-pooling operation along the channel dimension of the features, so that spatial information can be effectively preserved.

\noindent{\bfseries Analyses of LSC and SSM. }
The results from Figures~\ref{lsc} and ~\ref{ssm} show two aspects of observations as follows. Firstly, for input image pairs of the same sub-class, semantic information for distinguishing fine-grained objects can be enhanced, which ensures the final good results of object-aware spatial alignment (shown in Figure~\ref{ssm}). Secondly, for input image pairs of different sub-classes, our LSC suppresses the high response areas of the support features. As a result, the inter-class feature discrepancies will become much larger, and the query image will be classified as a different subclass from the support image. Besides, Figure~\ref{ssm} shows the features after applying the LSC and SSM, respectively, to further illustrate the alignment performance of our network in different stages.

\noindent{\bfseries Analyses of semantic correlation matrix in LSC. }
We conduct further studies from two aspects to demonstrate the effectiveness of the proposed semantic correlation matrix in LSC. The first aspect is to show whether high response values in the semantic correlation matrix can represent the semantically matched regions between the support and query images. The second aspect is to show whether the support features spatially transformed by the LSC are successfully matched with the query feature.

The first aspect can be well demonstrated by the semantic correlation matrix (1) in Figure~\ref{matrix}. The high response regions in the matrix indicate that the corresponding regions (cyan and orange regions) in the support and query features are highly correlated with each other.
The second aspect can be well demonstrated by the matrices (2) and (4) in Figure~\ref{matrix}, which represent the correlation matrices between the transformed support features and the query features. Note that higher response values on the diagonal of the correlation matrix imply stronger correlation between the transformed support features and the query features. It can be seen that the top left of the diagonal of the correlation matrix (2) are highly correlated and more concentrated on the diagonal, and closer to a symmetric matrix than those in the previous stage, which shows the transformed support features have good spatial match with the query features. In contrast, many rows of the matrix (4) are suppressed due to its semantic differences, which is also reflected on the diagonal of the correlation matrix.

\section{Conclusion}
We have introduced an object-aware long-short-range spatial alignment approach to boost the performance of few-shot fine-grained image classification. A foreground object enhancement module is first developed to strengthen those discriminative object features. Then a long-range semantic correspondence module is proposed to find a transferable feature transform for the enhanced features, to do the global support-query spatial matching. Finally, a short-range spatial manipulation module is developed to further refine the long-range transformed output by aligning more short-range local parts in the object. 
Experimental results on four benchmarks demonstrate the effectiveness and superiority of our proposed method. 

\section{Acknowledgement}
This work is supported by National Natural Science Foundation of China (No. 62071127 and U1909207), Shanghai Pujiang Program (No.19PJ1402000), Shanghai Municipal Science and Technology Major Project (No.2021SHZDZX0103), Shanghai Engineering Research Center of AI Robotics, Ministry of Education in China.

{\small
\bibliographystyle{ACM-Reference-Format}
\bibliography{ours.bbl}
}

\end{document}